# Deep Learning in Agriculture: A Survey

Andreas Kamilaris[1] and Francesc X. Prenafeta-Boldú

Institute for Food and Agricultural Research and Technology (IRTA)

**Abstract:** Deep learning constitutes a recent, modern technique for image processing and data analysis, with promising results and large potential. As deep learning has been successfully applied in various domains, it has recently entered also the domain of agriculture. In this paper, we perform a survey of 40 research efforts that employ deep learning techniques, applied to various agricultural and food production challenges. We examine the particular agricultural problems under study, the specific models and frameworks employed, the sources, nature and pre-processing of data used, and the overall performance achieved according to the metrics used at each work under study. Moreover, we study comparisons of deep learning with other existing popular techniques, in respect to differences in classification or regression performance. Our findings indicate that deep learning provides high accuracy, outperforming existing commonly used image processing techniques.

---

[1] Corresponding Author. Email: andreas.kamilaris@irta.cat

## 1. Introduction

Smart farming (Tyagi, 2016) is important for tackling the challenges of agricultural production in terms of productivity, environmental impact, food security and sustainability (Gebbers & Adamchuk, 2010). As the global population has been continuously increasing (Kitzes, et al., 2008), a large increase on food production must be achieved (FAO, 2009), maintaining at the same time availability and high nutritional quality across the globe, protecting the natural ecosystems by using sustainable farming procedures.

To address these challenges, the complex, multivariate and unpredictable agricultural ecosystems need to be better understood by monitoring, measuring and analyzing continuously various physical aspects and phenomena. This implies analysis of big agricultural data (Kamilaris, Kartakoullis, & Prenafeta-Boldú, A review on the practice of big data analysis in agriculture, 2017), and the use of new information and communication technologies (ICT) (Kamilaris, Gao, Prenafeta-Boldú, & Ali, 2016), both for short-scale crop/farm management as well as for larger-scale ecosystems' observation, enhancing the existing tasks of management and decision/policy making by context, situation and location awareness. Larger-scale observation is facilitated by remote sensing (Bastiaanssen, Molden, & Makin, 2000), performed by means of satellites, airplanes and unmanned aerial vehicles (UAV) (i.e. drones), providing wide-view snapshots of the agricultural environments. It has several advantages when applied to agriculture, being a well-known, non-destructive method to collect information about earth features while data may be obtained systematically over large geographical areas.

A large subset of the volume of data collected through remote sensing involve images. Images constitute, in many cases, a complete picture of the agricultural environments and could address a variety of challenges (Liaghat & Balasundram, 2010), (Ozdogan, Yang, Allez, & Cervantes, 2010). Hence, imaging analysis is an important research area in the agricultural domain and intelligent data analysis techniques are being used for image identification/classification, anomaly detection etc., in various agricultural applications (Teke, Deveci, Haliloğlu, Gürbüz, & Sakarya, 2013), (Saxena & Armstrong, 2014), (Singh, Ganapathysubramanian, Singh, & Sarkar, 2016). The most popular techniques and applications are presented in Appendix I, together with the sensing methods employed to acquire the images. From existing sensing methods, the most common one is satellite-based, using multi-spectral and hyperspectral imaging. Synthetic aperture radar (SAR), thermal and near infrared (NIR) cameras are being used in a lesser but increasing extent (Ishimwe, Abutaleb, & Ahmed, 2014), while optical and X-ray imaging are being applied in fruit and packaged food grading. The most popular techniques used for analyzing images include machine learning (ML) (K-means, support vector machines (SVM), artificial neural networks (ANN) amongst others), linear polarizations, wavelet-based filtering, vegetation indices (NDVI) and regression analysis (Saxena & Armstrong, 2014), (Singh, Ganapathysubramanian, Singh, & Sarkar, 2016).

Besides the aforementioned techniques, a new one which is recently gaining momentum is deep learning (DL) (LeCun, Bengio, & Hinton, 2015), (LeCun & Bengio, 1995). DL belongs to the machine learning computational field and is similar to ANN. However, DL is about "deeper" neural networks that provide a hierarchical representation of the data by means of various convolutions. This allows larger learning capabilities and thus higher performance and precision. A brief description of DL is attempted in Section 3.

The motivation for preparing this survey stems from the fact that DL in agriculture is a recent, modern and promising technique with growing popularity, while advancements and applications of DL in other domains indicate its large potential. The fact that today there exists at least 40 research efforts employing DL to address various agricultural problems with very good results, encouraged the authors to prepare this survey. To the authors' knowledge, this is the first such survey in the agricultural domain, while a small number of more general surveys do exist (Deng & Yu, 2014), (Wan, et al., 2014), (Najafabadi, et al., 2015), covering related work in DL in other domains.

## 2. Methodology

The bibliographic analysis in the domain under study involved two steps: a) collection of related work and b) detailed review and analysis of this work. In the first step, a keyword-based search for conference papers or journal articles was performed from the scientific databases IEEE Xplore and ScienceDirect, and from the web scientific indexing services Web of Science and Google Scholar. As search keywords, we used the following query:

[*"deep learning"*] AND [*"agriculture" OR "farming"*]

In this way, we filtered out papers referring to DL but not applied to the agricultural domain. From this effort, 47 papers had been initially identified. Restricting the search for papers with appropriate application of the DL technique and meaningful findings[2], the initial number of papers reduced to 40.

In the second step, the 40 papers selected from the previous step were analyzed one-by-one, considering the following research questions:

1. Which was the agricultural- or food-related problem they addressed?
2. Which was the general approach and type of DL-based models employed?
3. Which sources and types of data had been used?
4. Which were the classes and labels as modeled by the authors? Were there any variations among them, observed by the authors?
5. Any pre-processing of the data or data augmentation techniques used?
6. Which has been the overall performance depending on the metric adopted?
7. Did the authors test the performance of their models on different datasets?
8. Did the authors compare their approach with other techniques and, if yes, which was the difference in performance?

Our main findings are presented in Section 4 and the detailed information per paper is listed in Appendix II.

---

[2] A small number of papers claimed of using DL in some agricultural-related application, but they did not show any results nor provided performance metrics that could indicate the success of the technique used.

## 3. Deep Learning

DL extends classical ML by adding more "depth" (complexity) into the model as well as transforming the data using various functions that allow data representation in a hierarchical way, through several levels of abstraction (Schmidhuber, 2015), (LeCun & Bengio, 1995). A strong advantage of DL is feature learning, i.e. the automatic feature extraction from raw data, with features from higher levels of the hierarchy being formed by the composition of lower level features (LeCun, Bengio, & Hinton, 2015). DL can solve more complex problems particularly well and fast, because of more complex models used, which allow massive parallelization (Pan & Yang, 2010). These complex models employed in DL can increase classification accuracy or reduce error in regression problems, provided there are adequately large datasets available describing the problem. DL consists of various different components (e.g. convolutions, pooling layers, fully connected layers, gates, memory cells, activation functions, encode/decode schemes etc.), depending on the network architecture used (i.e. Unsupervised Pre-trained Networks, Convolutional Neural Networks, Recurrent Neural Networks, Recursive Neural Networks). The highly hierarchical structure and large learning capacity of DL models allow them to perform classification and predictions particularly well, being flexible and adaptable for a wide variety of highly complex (from a data analysis perspective) challenges (Pan & Yang, 2010). Although DL has met popularity in numerous applications dealing with raster-based data (e.g. video, images), it can be applied to any form of data, such as audio, speech, and natural language, or more generally to continuous or point data such as weather data (Sehgal, et al., 2017), soil chemistry (Song, et al., 2016) and population data (Demmers T. G., Cao, Parsons, Gauss, & Wathes, 2012). An example DL architecture is displayed in Figure 1, illustrating CaffeNet (Jia, et al., 2014), an example of a convolutional neural network, combining convolutional and fully connected (dense) layers.

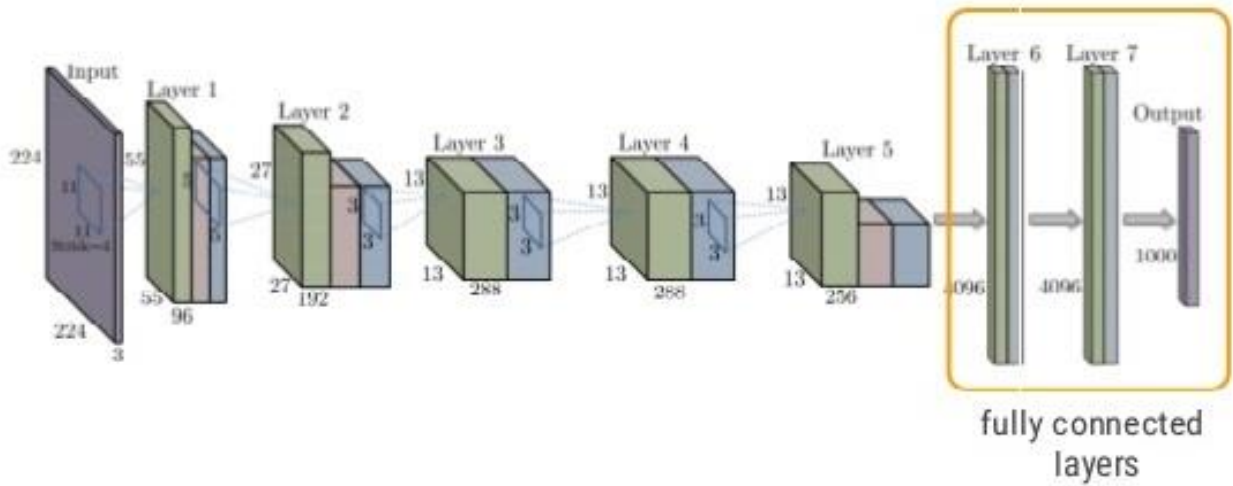

Figure 1: CaffeNet, an example CNN architecture. Source: *(Sladojevic, Arsenovic, Anderla, Culibrk, & Stefanovic, 2016)*

Convolutional Neural Networks (CNN) constitute a class of deep, feed-forward ANN, and they appear in numerous of the surveyed papers as the technique used (17 papers, 42%). As the figure shows, various convolutions are performed at some layers of the network, creating different representations of the learning dataset, starting from more general ones at the first larger layers, becoming more specific at the deeper layers. The convolutional layers act as feature extractors from the input images whose dimensionality is then reduced by the pooling layers. The convolutional layers encode multiple lower-level features into more discriminative features, in a way that is spatially context-aware. They may be understood as banks of filters that transform an input image into another, highlighting specific patterns. The fully connected layers, placed in many cases near the output of the model, act as classifiers exploiting the high-level features learned to classify input images in predefined classes or to make numerical predictions. They take a vector as input and produce another vector as output. An example visualization of leaf images after each processing step of the CaffeNet CNN, at a problem of identifying plant diseases, is depicted in Figure 2. We can observe that after each processing step, the particular elements of the image that reveal the indication of a disease become more evident, especially at the final step (Pool5).

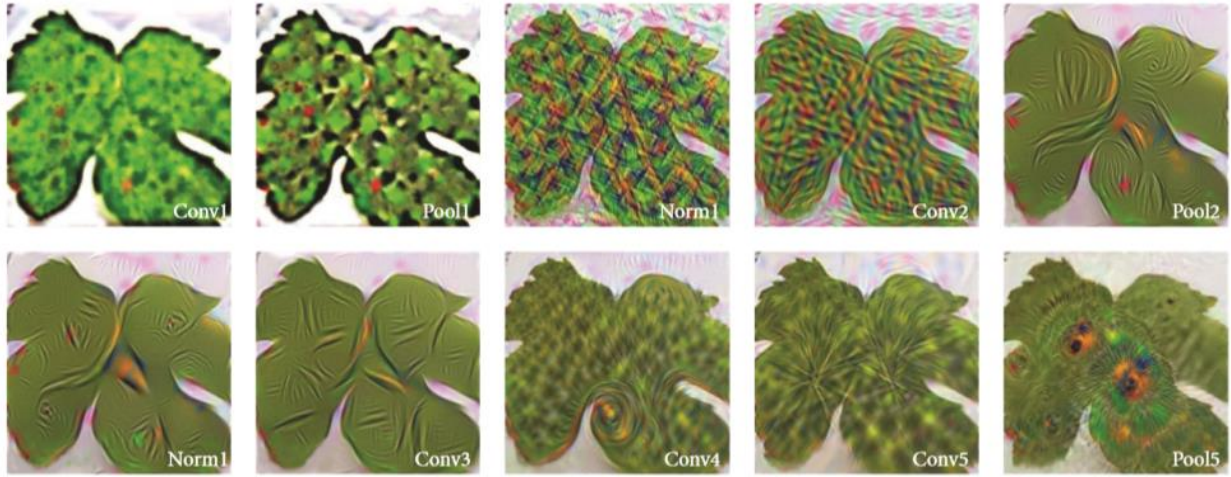

Figure 2: Visualization of the output layers images after each processing step of the CaffeNet CNN (i.e. convolution, pooling, normalization) at a plant disease identification problem based on leaf images. Source: *(Sladojevic, Arsenovic, Anderla, Culibrk, & Stefanovic, 2016)*

One of the most important advantages of using DL in image processing is the reduced need of feature engineering (FE). Previously, traditional approaches for image classification tasks had been based on hand-engineered features, whose performance affected heavily the overall results. FE is a complex, time-consuming process which needs to be altered whenever the problem or the dataset changes. Thus, FE constitutes an expensive effort that depends on experts' knowledge and does not generalize well (Amara, Bouaziz, & Algergawy, 2017). On the other hand, DL does not require FE, locating the important features itself through training.

A disadvantage of DL is the generally longer training time. However, testing time is generally faster than other methods ML-based methods (Chen, Lin, Zhao, Wang, & Gu, 2014). Other disadvantages include problems that might occur when using pre-trained models on datasets that are small or significantly different, optimization issues because of the models' complexity, as well as hardware restrictions.

In Section 5, we discuss over advantages and disadvantages of DL as they reveal through the surveyed papers.

## 3.1 Available Architectures, Datasets and Tools

There exist various successful and popular architectures, which researchers may use to start building their models instead of starting from scratch. These include AlexNet (Krizhevsky, Sutskever, & Hinton, 2012), CaffeNet (Jia, et al., 2014) (displayed in Figure 1), VGG (Simonyan & Zisserman, 2014), GoogleNet (Szegedy, et al., 2015) and Inception-ResNet (Szegedy, Ioffe, Vanhoucke, & Alemi, 2017), among others. Each architecture has different advantages and scenarios where it is more appropriate to be used (Canziani, Paszke, & Culurciello, 2016). It is also worth noting that almost all of the aforementioned models come along with their weights pre-trained, which means that their network had been already trained by some dataset and has thus learned to provide accurate classification for some particular problem domain (Pan & Yang, 2010). Common datasets used for pre-training DL architectures include ImageNet (Deng, et al., 2009) and PASCAL VOC (PASCAL VOC Project, 2012) (see also Appendix III).

Moreover, there exist various tools and platforms allowing researchers to experiment with DL (Bahrampour, Ramakrishnan, Schott, & Shah, 2015). The most popular ones are Theano, TensorFlow, Keras (which is an application programmer's interface on top of Theano and TensorFlow), Caffe, PyTorch, TFLearn, Pylearn2 and the Deep Learning Matlab Toolbox. Some of these tools (i.e. Theano, Caffe) incorporate popular architectures such as the ones mentioned above (i.e. AlexNet, VGG, GoogleNet), either as libraries or classes. For a more elaborate description of the DL concept and its applications, the reader could refer to existing bibliography (Schmidhuber, 2015), (Deng & Yu, 2014), (Wan, et al., 2014), (Najafabadi, et al., 2015), (Canziani, Paszke, & Culurciello, 2016), (Bahrampour, Ramakrishnan, Schott, & Shah, 2015).

## 4. Deep Learning Applications in Agriculture

In Appendix II, we list the 40 identified relevant works, indicating the agricultural-related research area, the particular problem they address, DL models and architectures implemented, sources of data used, classes and labels of the data, data pre-processing and/or augmentation employed, overall performance achieved according to the metrics adopted, as well as comparisons with other techniques, wherever available.

## 4.1 Areas of Use

Sixteen areas have been identified in total, with the popular ones being identification of weeds (5 papers), land cover classification (4 papers), plant recognition (4 papers), fruits counting (4 papers) and crop type classification (4 papers).

It is remarkable that all papers, except from (Demmers T. G., et al., 2010), (Demmers T. G., Cao, Parsons, Gauss, & Wathes, 2012) and (Chen, Lin, Zhao, Wang, & Gu, 2014), were published during or after 2015, indicating how recent and modern this technique is, in the domain of agriculture. More precisely, from the remaining 37 papers, 15 papers have been published in 2017, 15 in 2016 and 7 in 2015.

The large majority of the papers deal with image classification and identification of areas of interest, including detection of obstacles (e.g. (Steen, Christiansen, Karstoft, & Jørgensen, 2016), (Christiansen, Nielsen, Steen, Jørgensen, & Karstoft, 2016)) and fruit counting (e.g. (Rahnemoonfar & Sheppard, 2017), (Sa, et al., 2016)). Some papers focus on predicting future parameters, such as corn yield (Kuwata & Shibasaki, 2015) soil moisture content at the field (Song, et al., 2016) and weather conditions (Sehgal, et al., 2017).

From another perspective, most papers (20) target crops, while few works consider issues such as weed detection (8 papers), land cover (4 papers), research on soil (2 papers), livestock agriculture (3 papers), obstacle detection (3 papers) and weather prediction (1 paper).

## 4.2 Data Sources

Observing the sources of data used to train the DL model at every paper, large datasets of images are mainly used, containing thousands of images in some cases, either real ones (e.g. (Mohanty, Hughes, & Salathé, 2016), (Reyes, Caicedo, & Camargo, 2015), (Dyrmann, Karstoft, & Midtiby, 2016 )), or synthetic produced by the authors (Rahnemoonfar & Sheppard, 2017), (Dyrmann, Mortensen, Midtiby, & Jørgensen, 2016). Some datasets originate from well-known and publicly-available datasets such as PlantVillage, LifeCLEF, MalayaKew, UC Merced and Flavia (see Appendix III), while others constitute sets of real images collected by the authors for their research needs (e.g. (Sladojevic, Arsenovic, Anderla, Culibrk, & Stefanovic, 2016), (Bargoti & Underwood, 2016), (Xinshao & Cheng, 2015), (Sørensen, Rasmussen, Nielsen, & Jørgensen, 2017)). Papers dealing with land cover, crop type classification and yield estimation, as well as some papers related to weed detection employ a smaller number of images (e.g. tens of images), produced by UAV (Lu, et al., 2017), (Rebetez, J., et al., 2016), (Milioto, Lottes, & Stachniss, 2017), airborne (Chen, Lin, Zhao, Wang, & Gu, 2014), (Luus, Salmon, van den Bergh, & Maharaj, 2015) or satellite-based remote sensing (Kussul, Lavreniuk, Skakun, & Shelestov, 2017), (Minh, et al., 2017), (Ienco, Gaetano, Dupaquier, & Maurel, 2017), (Rußwurm & Körner, 2017). A particular paper investigating segmentation of root and soil uses images from X-ray tomography (Douarre, Schielein, Frindel, Gerth, & Rousseau, 2016). Moreover, some papers use text data, collected either from repositories (Kuwata & Shibasaki, 2015), (Sehgal, et al., 2017) or field sensors (Song, et al., 2016), (Demmers T. G., et al., 2010), (Demmers T. G., Cao, Parsons, Gauss, & Wathes, 2012). In general, the more complicated the problem to be solved, the more data is required. For example, problems involving large number of classes to identify (Mohanty, Hughes, & Salathé, 2016), (Reyes, Caicedo, & Camargo, 2015), (Xinshao & Cheng, 2015) and/or small Variation among the classes (Luus, Salmon, van den Bergh, & Maharaj, 2015), (Rußwurm & Körner, 2017), (Yalcin, 2017 ), (Namin, Esmaeilzadeh, Najafi, Brown, & Borevitz, 2017), (Xinshao & Cheng, 2015), require large number of input images to train their models.

**4.3 Data Variation**

Variation between classes is necessary for the DL models to be able to differentiate features and characteristics, and perform accurate classifications[3]. Hence, accuracy is positively correlated with variation among classes. Nineteen papers (47%) revealed some aspects of poor data variation. Luus et al. (Luus, Salmon, van den Bergh, & Maharaj, 2015) observed high relevance between some land cover classes (i.e. medium density and dense residential, buildings and storage tanks) while Ienco et al. (Ienco, Gaetano, Dupaquier, & Maurel, 2017) found that tree crops, summer crops and truck farming were classes highly mixed. A confusion between maize and soybeans was evident in (Kussul, Lavreniuk, Skakun, & Shelestov, 2017) and variation was low in botanically related crops, such as meadow, fallow, triticale, wheat, and rye (Rußwurm & Körner, 2017). Moreover, some particular views of the plants (i.e. flowers and leaf scans) offer different classification accuracy than branches, stems and photos of the entire plant. A serious issue in plant phenology recognition is the fact that appearances change very gradually and it is challenging to distinguish images falling into the growing durations that are in the middle of two successive stages (Yalcin, 2017 ), (Namin, Esmaeilzadeh, Najafi, Brown, & Borevitz, 2017). A similar issue appears when assessing the quality of vegetative development (Minh, et al., 2017). Furthermore, in the challenging problem of fruit counting, the models suffer from high occlusion, depth variation, and uncontrolled illumination, including high color similarity between fruit/foliage (Chen, et al., 2017), (Bargoti & Underwood, 2016). Finally, identification of weeds faces issues with respect to lighting, resolution, and soil type, and small variation between weeds and crops in shape, texture, color and position (i.e. overlapping) (Dyrmann, Karstoft, & Midtiby, 2016 ), (Xinshao & Cheng, 2015), (Dyrmann, Jørgensen, & Midtiby, 2017). In the large majority of the papers mentioned above (except from (Minh, et al., 2017)), this low variation has affected classification

---

[3] Classification accuracy is defined in Section 4.7 and Table 1.

accuracy significantly, i.e. more than 5%.

## 4.4 Data Pre-Processing

The large majority of related work (36 papers, 90%) involved some image pre-processing steps, before the image or particular characteristics/features/statistics of the image were fed as an input to the DL model. The most common pre-processing procedure was image resize (16 papers), in most cases to a smaller size, in order to adapt to the requirements of the DL model. Sizes of 256x256, 128x128, 96x96 and 60x60 pixels were common. Image segmentation was also a popular practice (12 papers), either to increase the size of the dataset (Ienco, Gaetano, Dupaquier, & Maurel, 2017), (Rebetez, J., et al., 2016), (Yalcin, 2017 ) or to facilitate the learning process by highlighting regions of interest (Sladojevic, Arsenovic, Anderla, Culibrk, & Stefanovic, 2016), (Mohanty, Hughes, & Salathé, 2016), (Grinblat, Uzal, Larese, & Granitto, 2016), (Sa, et al., 2016), (Dyrmann, Karstoft, & Midtiby, 2016 ), (Potena, Nardi, & Pretto, 2016) or to enable easier data annotation by experts and volunteers (Chen, et al., 2017), (Bargoti & Underwood, 2016). Background removal (Mohanty, Hughes, & Salathé, 2016), (McCool, Perez, & Upcroft, 2017), (Milioto, Lottes, & Stachniss, 2017), foreground pixel extraction (Lee, Chan, Wilkin, & Remagnino, 2015) or non-green pixels removal based on NDVI masks (Dyrmann, Karstoft, & Midtiby, 2016 ), (Potena, Nardi, & Pretto, 2016) were also performed to reduce the datasets' overall noise. Other operations involved the creation of bounding boxes (Chen, et al., 2017), (Sa, et al., 2016), (McCool, Perez, & Upcroft, 2017), (Milioto, Lottes, & Stachniss, 2017) to facilitate detection of weeds or counting of fruits. Some datasets were converted to grayscale (Santoni, Sensuse, Arymurthy, & Fanany, 2015), (Amara, Bouaziz, & Algergawy, 2017) or to the HSV color model (Luus, Salmon, van den Bergh, & Maharaj, 2015), (Lee, Chan, Wilkin, & Remagnino, 2015).

Furthermore, some papers used features extracted from the images as input to their models, such as shape and statistical features (Hall, McCool, Dayoub, Sunderhauf, &

Upcroft, 2015), histograms (Hall, McCool, Dayoub, Sunderhauf, & Upcroft, 2015), (Xinshao & Cheng, 2015), (Rebetez, J., et al., 2016), Principal Component Analysis (PCA) filters (Xinshao & Cheng, 2015), Wavelet transformations (Kuwata & Shibasaki, 2015) and Gray Level Co-occurrence Matrix (GLCM) features (Santoni, Sensuse, Arymurthy, & Fanany, 2015). Satellite or aerial images involved a combination of pre-processing steps such as orthorectification (Lu, et al., 2017), (Minh, et al., 2017) calibration and terrain correction (Kussul, Lavreniuk, Skakun, & Shelestov, 2017), (Minh, et al., 2017) and atmospheric correction (Rußwurm & Körner, 2017).

## 4.5 Data Augmentation

It is worth-mentioning that some of the related work under study (15 papers, 37%) employed data augmentation techniques (Krizhevsky, Sutskever, & Hinton, 2012), to enlarge artificially their number of training images. This helps to improve the overall learning procedure and performance, and for generalization purposes, by means of feeding the model with varied data. This augmentation process is important for papers that possess only small datasets to train their DL models, such as (Bargoti & Underwood, 2016), (Sladojevic, Arsenovic, Anderla, Culibrk, & Stefanovic, 2016), (Sørensen, Rasmussen, Nielsen, & Jørgensen, 2017), (Mortensen, Dyrmann, Karstoft, Jørgensen, & Gislum, 2016), (Namin, Esmaeilzadeh, Najafi, Brown, & Borevitz, 2017) and (Chen, et al., 2017). This process was especially important in papers where the authors trained their models using synthetic images and tested them on real ones (Rahnemoonfar & Sheppard, 2017) and (Dyrmann, Mortensen, Midtiby, & Jørgensen, 2016). In this case, data augmentation allowed their models to generalize and be able to adapt to the real-world problems more easily.

Transformations are label-preserving, and included rotations (12 papers), dataset partitioning/cropping (3 papers), scaling (3 papers), transposing (Sørensen, Rasmussen, Nielsen, & Jørgensen, 2017), mirroring (Dyrmann, Karstoft, & Midtiby, 2016 ), translations and perspective transform (Sladojevic, Arsenovic, Anderla, Culibrk, & Stefanovic, 2016), adaptations of objects' intensity in an object detection problem (Steen, Christiansen, Karstoft, & Jørgensen, 2016) and a PCA augmentation technique (Bargoti & Underwood, 2016).

Papers involving simulated data performed additional augmentation techniques such as varying the HSV channels and adding random shadows (Dyrmann, Mortensen, Midtiby, & Jørgensen, 2016) or adding simulated roots to soil images (Douarre, Schielein, Frindel, Gerth, & Rousseau, 2016).

## 4.6 Technical Details

From a technical side, almost half of the research works (17 papers, 42%) employed popular CNN architectures such as AlexNet, VGG16 and Inception-ResNet. From the rest, 14 papers developed their own CNN models, 2 papers adopted a first-order Differential Recurrent Neural Networks (DRNN) model, 5 papers preferred to use a Long Short-Term Memory (LSTM) model (Gers, Schmidhuber, & Cummins, 2000), one paper used deep belief networks (DBN) and one paper employed a hybrid of PCA with auto-encoders (AE). Some of the CNN approaches combined their model with a classifier at the output layer, such as logistic regression (Chen, Lin, Zhao, Wang, & Gu, 2014), Scalable Vector Machines (SVM) (Douarre, Schielein, Frindel, Gerth, & Rousseau, 2016), linear regression (Chen, et al., 2017), Large Margin Classifiers (LCM) (Xinshao & Cheng, 2015) and macroscopic cellular automata (Song, et al., 2016).

Regarding the frameworks used, all the works that employed some well-known CNN architecture had also used a DL framework, with Caffe being the most popular (13 papers, 32%), followed by Tensor Flow (2 papers) and deeplearning4j (1 paper). Ten research works developed their own software, while some authors decided to build their own models on top of Caffe (5 papers), Keras/Theano (5 papers), Keras/TensorFlow (4 papers), Pylearn2 (1 paper), MatConvNet (1 paper) and Deep Learning Matlab Toolbox (1 paper). A possible reason for the wide use of Caffe is that it incorporates various CNN frameworks and datasets, which can be used then easily and automatically by its users. Most of the studies divided their dataset between training and testing/verification data using a ratio of 80-20 or 90-10 respectively. In addition, various learning rates have been recorded, from 0.001 (Amara, Bouaziz, & Algergawy, 2017) and 0.005 (Mohanty, Hughes, & Salathé, 2016) up to 0.01 (Grinblat, Uzal, Larese, & Granitto, 2016). Learning rate is about how quickly a network learns. Higher values help avoid the solver being stuck in local minima, which can reduce performance significantly. A general approach used by many of the evaluated papers is to start out with a high learning rate and lower it as the training goes on. We note that learning rate is very dependent on the network architecture. Moreover, most of the research works that incorporated popular DL architectures took advantage of transfer learning (Pan & Yang, 2010), which concerns leveraging the already existing knowledge of some related task or domain in order to increase the learning efficiency of the problem under study by fine-tuning pre-trained models. As sometimes it is not possible to train a network from scratch due to having a small training data set or having a complex multi-task network, it is required that the network be at least partially initialized with weights from another pre-trained model. A common transfer learning technique is the use of pre-trained CNN, which are CNN models that have been already trained on some relevant dataset with possibly different number of classes. These models are then adapted to the particular challenge and dataset. This method was followed (among others) in (Lu, et al., 2017), (Douarre, Schielein, Frindel, Gerth, & Rousseau, 2016), (Reyes, Caicedo, & Camargo, 2015), (Bargoti & Underwood, 2016), (Steen, Christiansen, Karstoft, & Jørgensen, 2016), (Lee, Chan, Wilkin, & Remagnino, 2015), (Sa, et al., 2016), (Mohanty, Hughes, & Salathé, 2016), (Christiansen, Nielsen, Steen, Jørgensen, & Karstoft, 2016), (Sørensen, Rasmussen, Nielsen, & Jørgensen, 2017), for the VGG16, DenseNet, AlexNet and GoogleNet architectures.

## 4.7 Outputs

Finally, concerning the 31 papers that involved classification, the classes as used by the authors ranged from 2 (Lu, et al., 2017), (Pound, M. P., et al., 2016), (Douarre, Schielein, Frindel, Gerth, & Rousseau, 2016), (Milioto, Lottes, & Stachniss, 2017) up to 1,000 (Reyes, Caicedo, & Camargo, 2015). A large number of classes was observed in (Luus, Salmon, van den Bergh, & Maharaj, 2015) (21 land-use classes), (Rebetez, J., et al., 2016) (22 different crops plus soil), (Lee, Chan, Wilkin, & Remagnino, 2015) (44 plant species) and (Xinshao & Cheng, 2015) (91 classes of common weeds found in agricultural fields). In these papers, the number of outputs of the model was equal to the number of classes respectively. Each output was a different probability for the input image, segment, blob or pixel to belong to each class, and then the model picked the highest probability as its predicted class.

From the rest 9 papers, 2 performed predictions of fruits counted (scalar value as output) (Rahnemoonfar & Sheppard, 2017), (Chen, et al., 2017), 2 identified regions of fruits in the image (multiple bounding boxes) (Bargoti & Underwood, 2016), (Sa, et al., 2016), 2 predicted animal growth (scalar value) (Demmers T. G., et al., 2010), (Demmers T. G., Cao, Parsons, Gauss, & Wathes, 2012), one predicted weather conditions (scalar value) (Sehgal, et al., 2017), one crop yield index (scalar value) (Kuwata & Shibasaki, 2015) and one paper predicted percentage of soil moisture content (scalar value) (Song, et al., 2016).

## 4.8 Performance Metrics

Regarding methods used to evaluate performance, various metrics have been employed by the authors, each being specific to the model used at each study. Table 1 lists these metrics, together with their definition/description, and the symbol we use to refer to them in this survey. In some papers where the authors referred to accuracy without specifying its definition, we assumed they referred to classification accuracy (CA, first metric listed in Table 1). From this point onwards, we refer to "DL performance" as its score in some performance metric from the ones listed in Table 1.

Table 1: Performance metrics used in related work under study.

| No. | Performance Metric | Symbol Used | Description |
|---|---|---|---|
| 1. | Classification Accuracy | CA | The percentage of correct predictions where the top class (the one having the highest probability), as indicated by the DL model, is the same as the target label as annotated beforehand by the authors. For multi-class classification problems, CA is averaged among all the classes. CA is mentioned as Rank-1 identification rate in (Hall, McCool, Dayoub, Sunderhauf, & Upcroft, 2015). |
| 2. | Precision | P | The fraction of true positives (TP, correct predictions) from the total amount of relevant results, i.e. the sum of TP and false positives (FP). For multi-class classification problems, P is averaged among the classes. P=TP/(TP+FP) |
| 3. | Recall | R | The fraction of TP from the total amount of TP and false negatives (FN). For multi-class classification problems, R gets averaged among all the classes. R=TP/(TP+FN) |
| 4. | F1 score | F1 | The harmonic mean of precision and recall. For multi-class classification problems, F1 gets averaged among all the classes. It is mentioned as F-measure in (Minh, et al., 2017). F1=2 * (TP*FP) / (TP+FP) |
| 5. | LifeCLEF metric | LC | A score[4] related to the rank of the correct species in the list of retrieved species |
| 6. | Quality Measure | QM | Obtained by multiplying sensitivity (proportion of pixels that were detected correctly) and specificity (which proportion of detected pixels are truly correct). QM=TP2 / ((TP+FP)(TP+FN)) |
| 7. | Mean Square Error | MSE | Mean of the square of the errors between predicted and observed values. |
| 8. | Root Mean Square Error | RMSE | Standard deviation of the differences between predicted values and observed values. A normalized RMSE (N-RMSE) has been used in (Sehgal, et al., 2017). |
| 9. | Mean Relative Error | MRE | The mean error between predicted and observed values, in percentage. |
| 10. | Ratio of total fruits counted | RFC | Ratio of the predicted count of fruits by the model, with the actual count. The actual count was attained by taking the average count of individuals (i.e. experts or volunteers) observing the images independently. |
| 11. | L2 error | L2 | Root of the squares of the sums of the differences between predicted counts of fruits by the model and the actual counts. |
| 12. | Intersection over Union | IoU | A metric that evaluates predicted bounding boxes, by dividing the area of overlap between the predicted and the ground truth boxes, by the area of their union. An average (Dyrmann, Mortensen, Midtiby, & Jørgensen, 2016) or frequency weighted (Mortensen, Dyrmann, Karstoft, Jørgensen, & Gislum, 2016) IoU can be calculated. |
| 13. | CA-IoU, F1-IoU, | CA-IoU | These are the same CA, F1, P and R metrics as defined above, |

---

| | P-IoU or R-IoU | F1-IoU P-IoU R-IoU | combined with IoU in order to consider true/false positives/negatives. Used in problems involving bounding boxes. This is done by putting a minimum threshold on IoU, i.e. any value above this threshold would be considered as positive by the metric (and the model involved). Thresholds of 20% (Bargoti & Underwood, 2016), 40% (Sa, et al., 2016) and 50% (Steen, Christiansen, Karstoft, & Jørgensen, 2016), (Christiansen, Nielsen, Steen, Jørgensen, & Karstoft, 2016), (Dyrmann, Jørgensen, & Midtiby, 2017) have been observed[5]. |

CA was the most popular metric used (24 papers, 60%), followed by F1 (10 papers, 25%). Some papers included RMSE (4 papers), IoU (3 papers), RFC (Chen, et al., 2017), (Rahnemoonfar & Sheppard, 2017) or others. Some works used a combination of metrics to evaluate their efforts. We note that some papers employing CA, F1, P or R, used IoU in order to consider a model's prediction (Bargoti & Underwood, 2016), (Sa, et al., 2016), (Steen, Christiansen, Karstoft, & Jørgensen, 2016), (Christiansen, Nielsen, Steen, Jørgensen, & Karstoft, 2016), (Dyrmann, Jørgensen, & Midtiby, 2017). In these cases, a minimum threshold was put on IoU, and any value above this threshold would be considered as positive by the model.

We note that in some cases, a trade-off can exist between metrics. For example, in a weed detection problem (Milioto, Lottes, & Stachniss, 2017), it might be desirable to have a high R to eliminate most weeds, but not eliminating crops is of a critical importance, hence a lower P might be acceptable.

**4.9 Overall Performance**

We note that it is difficult if not impossible to compare between papers, as different metrics are employed for different tasks, considering different models, datasets and parameters. Hence, the reader should consider our comments in this section with some caution.

In 19 out of the 24 papers that involved CA as a metric, accuracy was high (i.e. above 90%), indicating good performance. The highest CA has been observed in the works of (Hall, McCool, Dayoub, Sunderhauf, & Upcroft, 2015), (Pound, M. P., et al., 2016), (Chen,

---

[5] In Appendix II, where we list the values of the metrics used at each paper, we denote CA-IoU(x), F1-IoU(x), P-IoU(x) or R-IoU(x), where x is the threshold (in percentage), over which results are considered as positive by the DL model employed.

Lin, Zhao, Wang, & Gu, 2014), (Lee, Chan, Wilkin, & Remagnino, 2015), (Minh, et al., 2017), (Potena, Nardi, & Pretto, 2016) and (Steen, Christiansen, Karstoft, & Jørgensen, 2016), with values of 98% or more, constituting remarkable results. From the 10 papers using F1 as metric, 5 had values higher than 0.90 with the highest F1 observed in (Mohanty, Hughes, & Salathé, 2016) and (Minh, et al., 2017) with values higher than 0.99. The works of (Dyrmann, Karstoft, & Midtiby, 2016 ), (Rußwurm & Körner, 2017), (Ienco, Gaetano, Dupaquier, & Maurel, 2017), (Mortensen, Dyrmann, Karstoft, Jørgensen, & Gislum, 2016), (Rebetez, J., et al., 2016), (Christiansen, Nielsen, Steen, Jørgensen, & Karstoft, 2016) and (Yalcin, 2017 ) were among the ones with the lowest CA (i.e. 73-79%) and/or F1 scores (i.e. 0.558 - 0.746), however state of the art work in these particular problems has shown lower CA (i.e. SVM, RF, Naïve- Bayes classifier). Particularly in (Rußwurm & Körner, 2017), the three-unit LSTM model employed provided 16.3% better CA than a CNN, which belongs to the family of DL. Besides, the above can be considered as "harder" problems, because of the use of satellite data (Ienco, Gaetano, Dupaquier, & Maurel, 2017), (Rußwurm & Körner, 2017) large number of classes (Dyrmann, Karstoft, & Midtiby, 2016 ), (Rußwurm & Körner, 2017), (Rebetez, J., et al., 2016), small training datasets (Mortensen, Dyrmann, Karstoft, Jørgensen, & Gislum, 2016), (Christiansen, Nielsen, Steen, Jørgensen, & Karstoft, 2016) or very low variation among the classes (Yalcin, 2017 ), (Dyrmann, Karstoft, & Midtiby, 2016 ), (Rebetez, J., et al., 2016).

**4.10 Generalizations on Different Datasets**

It is important to examine whether the authors had tested their implementations on the same dataset (e.g. by dividing the dataset into training and testing/validation sets) or used different datasets to test their solution. From the 40 papers, only 8 (20%) used different datasets for testing than the one for training. From these, 2 approaches trained their models by using simulated data and tested on real data (Dyrmann, Mortensen, Midtiby, & Jørgensen, 2016), (Rahnemoonfar & Sheppard, 2017) and 2 papers tested their models on a dataset produced 2-4 weeks after, with a more advanced growth stage of plants and weeds (Milioto, Lottes, & Stachniss, 2017), (Potena, Nardi, & Pretto, 2016). Moreover, 3 papers used different fields for testing than the ones used for training (McCool, Perez, & Upcroft, 2017), with a severe degree of occlusion compared to the other training field (Dyrmann, Jørgensen, & Midtiby, 2017), or containing other obstacles such as people and animals (Steen, Christiansen, Karstoft, & Jørgensen, 2016). Sa et al. (Sa, et al., 2016) used a different dataset to evaluate whether the model can generalize on different fruits. From the other 32 papers, different trees were used in training and testing in (Chen, et al., 2017), while different rooms for pigs (Demmers T. G., Cao, Parsons, Gauss, & Wathes, 2012) and chicken (Demmers T. G., et al., 2010) were considered. Moreover, Hall et al. applied condition variations in testing (i.e. translations, scaling, rotations, shading and occlusions) (Hall, McCool, Dayoub, Sunderhauf, & Upcroft, 2015) while scaling for a certain range translation distance and rotation angle was performed on the testing dataset in (Xinshao & Cheng, 2015). The rest 27 papers did not perform any changes between the training/testing datasets, a fact that lowers the overall confidence for the results presented. Finally, it is interesting to observe how these generalizations affected the performance of the models, at least in cases where both data from same and different datasets were used in testing. In (Sa, et al., 2016), F1-IoU(40) was higher for the detection of apples (0.938), strawberry (0.948), avocado (0.932) and mango (0.942), than in the default case of sweet pepper (0.838). In (Rahnemoonfar & Sheppard, 2017), RFC was 2% less in the real images than in the synthetic ones. In (Potena, Nardi, & Pretto, 2016), CA was 37.6% less at the dataset involving plants of 4-weeks more advanced growth. According to the authors, the model was trained based on plants that were in their first growth stage, thus without their complete morphological features, which were included in the testing dataset. Moreover, in (Milioto, Lottes, & Stachniss, 2017) P was 2% higher at the 2-weeks more advanced growth dataset, with 9% lower R.

Hence, in the first case there was improvement in performance (Sa, et al., 2016), and in the last three cases a reduction, slight one in (Rahnemoonfar & Sheppard, 2017) and (Milioto, Lottes, & Stachniss, 2017) but considerable in (Potena, Nardi, & Pretto, 2016). From the other papers using different testing datasets, as mentioned above, high percentages of CA (94-97.3%), P-IoU (86.6%) and low values of MRE (1.8 -10%) have been reported. These show that the DL models were able to generalize well to different datasets. However, without more comparisons, this is only a speculation that can be figured out of the small number of observations available.

## 4.11 Performance Comparison with Other Approaches

A critical aspect of this survey is to examine how DL performs in relation to other existing techniques. The 14[th] column of Appendix II presents whether the authors of related work compared their DL-based approach with other techniques used for solving their problem under study. We focus only on comparisons between techniques used for the same dataset at the same research paper, with the same metric.

In almost all cases, the DL models outperform other approaches implemented for comparison purposes. CNN show 1-8% higher CA in comparison to SVM (Chen, Lin, Zhao, Wang, & Gu, 2014), (Lee, Chan, Wilkin, & Remagnino, 2015), (Grinblat, Uzal, Larese, & Granitto, 2016), (Pound, M. P., et al., 2016), 41% improvement of CA when compared to ANN (Lee, Chan, Wilkin, & Remagnino, 2015) and 3-8% higher CA when compared to RF (Kussul, Lavreniuk, Skakun, & Shelestov, 2017), (Minh, et al., 2017), (McCool, Perez, & Upcroft, 2017), (Potena, Nardi, & Pretto, 2016), (Hall, McCool, Dayoub, Sunderhauf, & Upcroft, 2015). CNN also seem to be superior than unsupervised feature learning with 3-11% higher CA (Luus, Salmon, van den Bergh, & Maharaj, 2015), 2-44% improved CA in relation to local shape and color features (Dyrmann, Karstoft, & Midtiby, 2016 ), (Sørensen, Rasmussen, Nielsen, & Jørgensen, 2017), and 2% better CA (Kussul, Lavreniuk, Skakun, & Shelestov, 2017) or 18% less RMSE (Song, et al., 2016) compared to multilayer perceptrons. CNN had also superior performance than Penalized Discriminant Analysis (Grinblat, Uzal, Larese, & Granitto, 2016), SVM Regression (Kuwata & Shibasaki, 2015), area-based techniques (Rahnemoonfar & Sheppard, 2017), texture-based regression models (Chen, et al., 2017), LMC classifiers (Xinshao & Cheng, 2015), Gaussian Mixture Models (Santoni, Sensuse, Arymurthy, & Fanany, 2015) and Naïve-Bayes classifiers (Yalcin, 2017 ).

In cases where Recurrent Neural Networks (RNN) (Mandic & Chambers, 2001) architectures were employed, the LSTM model had 1% higher CA than RF and SVM in (Ienco, Gaetano, Dupaquier, & Maurel, 2017), 44% improved CA than SVM in (Rußwurm & Körner, 2017) and 7-9% better CA than RF and SVM in (Minh, et al., 2017).

In only one case, DL showed worse performance against another technique, and this was when a CNN was compared to an approach involving local descriptors to represent images together with KNN as the classification strategy (20% worse LC) (Reyes, Caicedo, & Camargo, 2015).

## 5. Discussion

Our analysis has shown that DL offers superior performance in the vast majority of related work. When comparing the performance of DL-based approaches with other techniques at each paper, it is of paramount importance to adhere to the same experimental conditions (i.e. datasets and performance metrics). From the related work under study, 28 out of the 40 papers (70%) performed direct, valid and correct comparisons among the DL-based approach employed and other state-of-art techniques used to solve the particular problem tackled at each paper. Due to the fact that each paper involved different datasets, pre-processing techniques, metrics, models and parameters, it is difficult if not impossible to generalize and perform comparisons between papers. Thus, our comparisons have been strictly limited among the techniques used at each paper. Thus, based on these constraints, we have observed that DL has outperformed traditional approaches used such as SVM, RF, ANN, LMC classifiers and others. It seems that the automatic feature extraction performed by DL models is more effective than the feature extraction process through traditional approaches such as Scale Invariant Feature Transform (SIFT), GLCM, histograms, area-based techniques (ABT), statistics-, texture-, color- and shape-based algorithms, conditional random fields to model color and visual texture features, local de-correlated channel features and other manual feature extraction techniques. This is reinforced by the combined CNN+LSTM model employed in (Namin, Esmaeilzadeh, Najafi, Brown, & Borevitz, 2017), which outperformed a LSTM model which used hand crafted feature descriptors as inputs by 25% higher CA. Interesting attempts to combine hand-crafted features and CNN-based features were performed in (Hall, McCool, Dayoub, Sunderhauf, & Upcroft, 2015) and (Rebetez, J., et al., 2016).

Although DL has been associated with computer vision and image analysis (which is also the general case in this survey), we have observed 5 related works where DL-based models have been trained based on field sensory data (Kuwata & Shibasaki, 2015), (Sehgal, et al., 2017) and a combination of static and dynamic environmental variables (Song, et al., 2016), (Demmers T. G., et al., 2010), (Demmers T. G., Cao, Parsons, Gauss, & Wathes, 2012). These papers indicate the potential of DL to be applied in a wide variety of agricultural problems, not only those involving images.

Examining agricultural areas where DL techniques have been applied, leaf classification, leaf and plant disease detection, plant recognition and fruit counting have some papers which present very good performance (i.e. CA > 95%, F1 > 0.92 or RFC > 0.9). This is probably because of the availability of datasets in these domains, as well as the distinct characteristics of (sick) leaves/plants and fruits in the image. On the other hand, some papers in land cover classification, crop type classification, plant phenology recognition and weed detection showed average performance (i.e. CA < 87% or F1 < 0.8). This could be due to leaf occlusion in weed detection, use of noise-prone satellite imagery in land cover problems, crops with low variation and botanical relationship or the fact that appearances change very gradually while plants grow in phenology recognition efforts. Without underestimating the quality of any of the surveyed papers, we highlight some that claim high performance (CA > 91%, F1-IoU(20) > 0.90 or RFC > 0.91), considering the complexity of the problem in terms of its definition or the large number of classes involved (more than 21 classes). These papers are the following: (Mohanty, Hughes, & Salathé, 2016), (Luus, Salmon, van den Bergh, & Maharaj, 2015), (Lee, Chan, Wilkin, & Remagnino, 2015), (Rahnemoonfar & Sheppard, 2017), (Chen, et al., 2017), (Bargoti & Underwood, 2016), (Xinshao & Cheng, 2015) and (Hall, McCool, Dayoub, Sunderhauf, & Upcroft, 2015). We also highlight papers that trained their models on simulated data, and tested them on real data, which are (Dyrmann, Mortensen, Midtiby, & Jørgensen, 2016), (Rahnemoonfar & Sheppard, 2017), and (Douarre, Schielein, Frindel, Gerth, & Rousseau, 2016). These works constitute important efforts in the DL community, as they attempt to solve the problem of inexistent or not large enough datasets in various problems. Finally, as discussed in Section 4.10, most authors used the same datasets for training and testing their implementation, a fact that lowers the confidence in the overall findings, although there have been indications that the models seem to generalize well, with only small reductions in performance.

## 5.1 Advanced Deep Learning Applications

Although the majority of papers used typical CNN architectures to perform classification (23 papers, 57%), some authors experimented with more advanced models in order to solve more complex problems, such as crop type classification from UAV imagery (CNN + HistNN using RGB histograms) (Rebetez, J., et al., 2016), estimating number of tomato fruits (Modified Inception-ResNet CNN) (Rahnemoonfar & Sheppard, 2017) and estimating number of orange or apple fruits (CNN adapted for blob detection and counting + Linear

Regression) (Chen, et al., 2017). Particularly interesting were the approaches employing the Faster Region-based CNN + VGG16 model (Bargoti & Underwood, 2016), (Sa, et al., 2016), in order not only to count fruits and vegetables, but also to locate their placement in the image by means of bounding boxes. Similarly, the work in (Dyrmann, Jørgensen, & Midtiby, 2017) used the DetectNet CNN to detect bounding boxes of weed instances in images of cereal fields. These approaches (Faster Region-based CNN, DetectNet CNN) constitute a very promising research direction, since the task of identifying the bounding box of fruits/vegetables/weeds in an image has numerous real-life applications and could solve various agricultural problems

Moreover, considering not only space but also time series, some authors employed RNN-based models in land cover classification (one-unit LSTM model + SVM) (Ienco, Gaetano, Dupaquier, & Maurel, 2017), crop type classification (three-unit LSTM) (Rußwurm & Körner, 2017), classification of different accessions of Arabidopsis thaliana based on successive top-view images (CNN+ LSTM) (Namin, Esmaeilzadeh, Najafi, Brown, & Borevitz, 2017), mapping winter vegetation quality coverage (Five-unit LSTM, Gated Recurrent Unit) (Minh, et al., 2017), estimating the weight of pigs or chickens (DRNN) (Demmers T. G., et al., 2010), (Demmers T. G., Cao, Parsons, Gauss, & Wathes, 2012) and for predicting weather based on previous year's conditions (LSTM) (Sehgal, et al., 2017). RNN-based models offer higher performance, as they can capture the time dimension, which is impossible to be exploited by simple CNN. RNN architectures tend to exhibit dynamic temporal behavior, being able to record long-short temporal dependencies, remembering and forgetting after some time or when needed (i.e. LSTM). Differences in performance between RNN and CNN are distinct in the related work under study, as shown in Table 2. This 16% improvement in CA could be attributed to the additional information provided by the time series. For example, in the crop type classification case (Rußwurm & Körner, 2017), the authors mention, "*crops change their* *spectral characteristics due to environmental influences and can thus not be monitored effectively with classical mono-temporal approaches. Performance of temporal models increases at the beginning of vegetation period*". LSTM-based approaches work well also for low represented and difficult classes, as demonstrated in (Ienco, Gaetano, Dupaquier, & Maurel, 2017).

Table 2: Difference in Performance between CNN and RNN.

| No. | Application in Agriculture | Performance Metric | Difference in Performance | Reference |
|---|---|---|---|---|
| 1. | Crop type classification considering time series | CA, F1 | Three-unit LSTM: 76.2% (CA), 0.558 (F1) CNN: 59.9% (CA), 0.236 (F1) | (Rußwurm & Körner, 2017) |
| 2. | Classify the phenotyping of Arabidopsis in four accessions | CA | CNN+ LSTM: 93% CNN: 76.8% | (Namin, Esmaeilzadeh, Najafi, Brown, & Borevitz, 2017) |

Finally, the critical aspect of fast processing of DL models in order to be easily used in robots for real-time decision making (e.g. detection of weeds) was examined in (McCool, Perez, & Upcroft, 2017), and it is worth-mentioning. The authors have showed that a lightweight implementation had only a small penalty in CA (3.90%), being much faster (i.e. processing of 40.6 times more pixels per second). They proposed the idea of "teacher and student networks", where the teacher is the more heavy approach that helps the student (light implementation) to learn faster and better.

## 5.2 Advantages of Deep Learning

Except from improvements in performance of the classification/prediction problems in the surveyed works (see Sections 4.9 and 4.11), the advantage of DL in terms of reduced effort in feature engineering was demonstrated in many of the papers. Hand-engineered components require considerable time, an effort that takes place automatically in DL. Besides, sometimes manual search for good feature extractors is not an easy and obvious task. For example, in the case of estimating crop yield (Kuwata & Shibasaki, 2015), extracting manually features that significantly affected crop growth was not possible. This was also the case of estimating the soil moisture content (Song, et al., 2016).

Moreover, DL models seem to generalize well. For example, in the case of fruit counting, the model learned explicitly to count (Rahnemoonfar & Sheppard, 2017). In the banana leaf classification problem (Amara, Bouaziz, & Algergawy, 2017), the model was robust under challenging conditions such as illumination, complex background, different resolution, size and orientation of the images. Also in the fruits counting papers (Chen, et al., 2017), (Rahnemoonfar & Sheppard, 2017), the models were robust to occlusion, variation, illumination and scale. The same detection frameworks could be used for a variety of circular fruits such as peaches, citrus, mangoes etc. As another example, a key feature of the DeepAnomaly model was the ability to detect unknown objects/anomalies and not just a set of predefined objects, exploiting the homogeneous characteristics of an agricultural field to detect distant, heavy occluded and unknown objects (Christiansen, Nielsen, Steen, Jørgensen, & Karstoft, 2016). Moreover, in the 8 papers mentioned in Section 4.10 where different datasets were used for testing, the performance of the model was generally high, with only small reductions in performance in comparison with the performance when using the same dataset for training and testing.

Although DL takes longer time to train than other traditional approaches (e.g. SVM, RF), its testing time efficiency is quite fast. For example, in detecting obstacles and anomaly (Christiansen, Nielsen, Steen, Jørgensen, & Karstoft, 2016), the model took much longer to train, but after it did, its testing time was less than the one of SVM and KNN. Besides, if we take into account the time needed to manually design filters and extract features, "*the time used on annotating images and training the CNN becomes almost negligible*" (Sørensen, Rasmussen, Nielsen, & Jørgensen, 2017).

Another advantage of DL is the possibility to develop simulated datasets to train the model, which could be properly designed in order to solve real-world problems. For example, in the issue of detecting weeds and maize in fields (Dyrmann, Mortensen,

Midtiby, & Jørgensen, 2016), the authors overcame the plant foliage overlapping problem by simulating top-down images of overlapping plants on soil background. The trained network was then capable of distinguish weeds from maize even in overlapping conditions.

## 5.3 Disadvantages and Limitations of Deep Learning

A considerable drawback and barrier in the use of DL is the need of large datasets, which would serve as the input during the training procedure. In spite of data augmentation techniques which augment some dataset with label-preserving transformations, in reality at least some hundreds of images are required, depending on the complexity of the problem under study (i.e. number of classes, precision required etc.). For example, the authors in (Mohanty, Hughes, & Salathé, 2016) and (Sa, et al., 2016) commented that a more diverse set of training data was needed to improve CA. A big problem with many datasets is the low variation among the different classes (Yalcin, 2017 ), as discussed in Section 4.3, or the existence of noise, in the form of low resolution, inaccuracy of sensory equipment (Song, et al., 2016), crops' occlusions, plants overlapping and clustering, and others.

As data annotation is a necessary operation in the large majority of cases, some tasks are more complex and there is a need for experts (who might be difficult to involve) in order to annotate input images. As mentioned in (Amara, Bouaziz, & Algergawy, 2017), there is a limited availability of resources and expertise on banana pathology worldwide. In some cases, experts or volunteers are susceptible to errors during data labeling, especially when this is a challenging task e.g. fruit count (Chen, et al., 2017), (Bargoti & Underwood, 2016) or to determine if images contain weeds or not (Sørensen, Rasmussen, Nielsen, & Jørgensen, 2017), (Dyrmann, Jørgensen, & Midtiby, 2017).

Another limitation is the fact that the DL models can learn some problem particularly well, even generalize in some aspects as mentioned in Section 5.2, but they cannot generalize beyond the "boundaries of the dataset's expressiveness". For example, classification of single leaves, facing up, on a homogeneous background is performed in (Mohanty,

Hughes, & Salathé, 2016). A real world application should be able to classify images of a disease as it presents itself directly on the plant. Many diseases do not present themselves on the upper side of leaves only. As another example, plant recognition in (Lee, Chan, Wilkin, & Remagnino, 2015) was noticeably affected by environmental factors such as wrinkled surface and insect damages. The model for counting tomatoes in (Rahnemoonfar & Sheppard, 2017) could count ripe and half-ripe fruits, however, "*it failed to count green fruits because it was not trained for this purpose*". If an object size in a testing image was significantly less than that of a training set, the model missed the detection in (Sa, et al., 2016). Difficulty in detecting heavily occluded and distant objects was observed in (Christiansen, Nielsen, Steen, Jørgensen, & Karstoft, 2016). Occlusion was a serious issue also in (Hall, McCool, Dayoub, Sunderhauf, & Upcroft, 2015).

A general issue in computer vision, not only in DL, is that data pre-processing is sometimes a necessary and time-consuming task, especially when satellite or aerial photos are involved, as we saw in Section 4.4. A problem with hyperspectral data is their high dimensionality and limited training samples (Chen, Lin, Zhao, Wang, & Gu, 2014). Moreover, sometimes the existing datasets do not describe completely the problem they target (Song, et al., 2016). As an example, for estimating corn yield (Kuwata & Shibasaki, 2015), it was necessary to consider also external factors other than the weather by inputting cultivation information such as fertilization and irrigation.

Finally, in the domain of agriculture, there do not exist many publicly available datasets for researchers to work with, and in many cases, researchers need to develop their own sets of images. This could require many hours or days of work.

## 5.4 Future of Deep Learning in Agriculture

Observing Appendix I, which lists various existing applications of computer vision in agriculture, we can see that only the problems of land cover classification, crop type estimation, crop phenology, weed detection and fruit grading have been approximated using DL. It is interesting to see how DL would behave also in other agricultural-related problems listed in Appendix I, such as seeds identification, soil and leaf nitrogen content, irrigation, plants' water stress detection, water erosion assessment, pest detection, herbicide use, identification of contaminants, diseases or defects on food, crop hail damage and greenhouse monitoring. Intuitively, since many of the aforementioned research areas employ data analysis techniques (see Appendix I) with similar concepts and comparable performance to DL (i.e. linear and logistic regression, SVM, KNN, K-means clustering, Wavelet-based filtering, Fourier transform) (Singh, Ganapathysubramanian, Singh, & Sarkar, 2016), then it could be worth to examine the applicability of DL on these problems too.

Other possible application areas could be the use of aerial imagery (i.e. by means of drones) to monitor the effectiveness of the seeding process, to increase the quality of wine production by harvesting grapes at the right moment for best maturity levels, to monitor animals and their movements to consider their overall welfare and identify possible diseases, and many other scenarios where computer vision is involved.

In spite of the limited availability of open datasets in agriculture, In Appendix III, we list some of the most popular, free to download datasets available on the web, which could be used by researchers to start testing their DL architectures. These datasets could be used to pre-train DL models and then adapt them to more specific future agricultural challenges. In addition to these datasets, remote sensing data containing multi-temporal, multi-spectral and multi-source images that could be used in problems related to land and crop cover classification are available from satellites such as MERIS, MODIS, AVHRR, RapidEye, Sentinel, Landsat etc.

More approaches adopting LSTM or other RNN models are expected in the future, exploiting the time dimension to perform higher performance classification or prediction. An example application could be to estimate the growth of plants, trees or even animals based on previous consecutive observations, to predict their yield, assess their water needs or avoid diseases from occurring. These models could find applicability in environmental informatics too, for understanding climatic change, predicting weather conditions and phenomena, estimating the environmental impact of various physical or artificial processes (Kamilaris, Assumpcio, Blasi, Torrellas, & Prenafeta-Boldú, 2017) etc. Related work under study involved up to a five-unit LSTM model (Minh, et al., 2017). We expect in the future to see more layers stacked together in order to build more complex LSTM architectures (Ienco, Gaetano, Dupaquier, & Maurel, 2017). We also believe that datasets with increasing temporal sequence length will appear, which could improve the performance of LSTM (Rußwurm & Körner, 2017).

Moreover, more complex architectures would appear, combining various DL models and classifiers together, or combining hand-crafted features with automatic features extracted by using various techniques, fused together to improve the overall outcome, similar to what performed in (Hall, McCool, Dayoub, Sunderhauf, & Upcroft, 2015) and (Rebetez, J., et al., 2016). Researchers are expected to test their models using more general and realistic dataset, demonstrating the ability of the models to generalize to various real-world situations. A combination of popular performance metrics, such as the ones mentioned in Table 1, are essential to be adopted by the authors for comparison purposes. It would be desirable if researchers made their datasets publicly available, for use also by the general research community.

Finally, some of the solutions discussed in the surveyed papers could have a commercial use in the near future. The approaches incorporating Faster Region-based CNN and DetectNet CNN (Bargoti & Underwood, 2016), (Chen, et al., 2017), (Rahnemoonfar & Sheppard, 2017) would be extremely useful for automatic robots that collect crops, remove weeds or for estimating the expected yields of various crops. A future application of this technique could be also in microbiology for human or animal cell counting (Chen, et al., 2017). The DRNN model controlling the daily feed intake of pigs or chicken, predicting quite accurately the required feed intake for the whole of the growing period, would be useful to farmers when deciding on a growth curve suitable for various scenarios. Following some growth patterns would have potential advantages for animal welfare in terms of leg health, without compromising the idea animals' final weight and total feed intake requirement (Demmers T. G., et al., 2010), (Demmers T. G., Cao, Parsons, Gauss, & Wathes, 2012).

## 6. Conclusion

In this paper, we have performed a survey of deep learning-based research efforts applied in the agricultural domain. We have identified 40 relevant papers, examining the particular area and problem they focus on, technical details of the models employed, sources of data used, pre-processing tasks and data augmentation techniques adopted, and overall performance according to the performance metrics employed by each paper. We have then compared deep learning with other existing techniques, in terms of performance. Our findings indicate that deep learning offers better performance and outperforms other popular image processing techniques. For future work, we plan to apply the general concepts and best practices of deep learning, as described through this survey, to other areas of agriculture where this modern technique has not yet been adequately used. Some of these areas have been identified in the discussion section.

Our aim is that this survey would motivate more researchers to experiment with deep learning, applying it for solving various agricultural problems involving classification or prediction, related to computer vision and image analysis, or more generally to data analysis. The overall benefits of deep learning are encouraging for its further use towards smarter, more sustainable farming and more secure food production.

**Acknowledgments**

We would like to thank the reviewers, whose valuable feedback, suggestions and comments increased significantly the overall quality of this survey. This research has been supported by the P-SPHERE project, which has received funding from the European Union's Horizon 2020 research and innovation programme under the Marie Skodowska-Curie grant agreement No 665919.

Appendix I: Applications of computer vision in agriculture and popular techniques used.

| No. | Application in Agriculture | Remote sensing | Techniques for data analysis |
|---|---|---|---|
| 1. | Soil and vegetation/crop mapping | Hyperspectral imaging (satellite and airborne), multi-spectral imaging (satellite), synthetic aperture radar (SAR) | Image fusion, SVM, end-member extraction algorithm, co-polarized phase differences (PPD), linear polarizations (HH, VV, HV), distance-based classification, decision trees, linear mixing models, logistic regression, ANN, NDVI |
| 2. | Leaf area index and crop canopy | Hyperspectral imaging (airborne), multi-spectral imaging (airborne) | Linear regression analysis, NDVI |
| 3. | Crop phenology | Satellite remote sensing (general) | Wavelet-based filtering, Fourier transforms, NDVI |
| 4. | Crop height, estimation of yields, fertilizers' effect and biomass | Light Detection and Ranging (LIDAR), hyperspectral and multi-spectral imaging, SAR, red-edge camera, thermal infrared | Linear and exponential regression analysis, linear polarizations (VV), wavelet-based filtering, vegetation indices (NDVI, ICWSI), ANN |
| 5. | Crop monitoring | Satellite remote sensing, (hyperspectral and multi-spectral imaging), NIR camera, SAR | Stepwise discriminate analysis (DISCRIM) feature extraction, linear regression analysis, co-polarized phase differences (PPD), linear polarizations (HH, VV, HV, RR and RL), classification and regression tree analysis |
| 6. | Identification of seeds and reorganization of species | Remote sensing in general, cameras and photo-detectors, hyperspectral imaging | Principal component analysis, feature extraction,  linear regression analysis |
| 7. | Soil and leaf nitrogen content and treatment, salinity detection | Hyperspectral and multi-spectral imaging, thermal imaging | Linear and exponential regression analysis |
| 8. | Irrigation | Satellite remote sensing (hyperspectral and multi-spectral imaging), red-edge camera, thermal infrared | Image classification techniques (unsupervised clustering, density slicing with thresholds), decision trees, linear regression analysis, NDVI |
| 9. | Plants water stress detection and drought conditions | Satellite remote sensing (hyperspectral and multi-spectral imaging, radar images), thermal imaging, NIR camera, red-edge camera | Fraunhofer Line Depth (FLD) principle,  linear regression analysis, NDVI |
| 10. | Water erosion assessment | Satellite remote sensing (optical and radar images), SAR, NIR camera | Interferometric SAR image processing,  linear and exponential regression analysis, contour tracing, linear polarizations (HH, VV) |
| 11. | Pest detection and management | Hyperspectral and multi-spectral imaging, microwave remote sensing, thermal camera | Image processing using sample imagery, linear and exponential regression analysis, statistical analysis, CEM nonlinear signal processing, NDVI |
| 12. | Weed detection | Remote sensing in general, optical cameras and photo-detectors, hyperspectral and multi-spectral imaging | Pixel classification based on k-means clustering and Bayes classifier, feature extraction techniques with FFT and GLCM, wavelet-based classification and Gabor filtering, genetic algorithms, fuzzy techniques, artificial neural networks, erosion and dilation segmentation, logistic regression, edge detection, color detection, principal component analysis |
| 13. | Herbicide | Remote sensing in general, optical cameras and photo-detectors | Fuzzy techniques, discriminant analysis |
| 14. | Fruit grading | Optical cameras and photo-detectors, monochrome images with different illuminations | K-means clustering, image fusion, color histogram techniques, machine learning (esp. SVM), Bayesian discriminant analysis, Bayes filtering, linear discriminant analysis |
| 15. | Packaged food and food products – identification of contaminants, diseases or defects, bruise detection | X-ray imaging (or transmitted light), CCD cameras, monochrome images with different illuminations, thermal cameras, multi-spectral and hyperspectral NIR-based imaging | 3D vision, invariance, pattern recognition and image modality, multivariate image analysis with principal component analysis, K-mean clustering, SVM, linear discriminant analysis, classification trees, K-nearest neighbors, decision trees, fusion, feature extraction techniques with FFT, standard Bayesian discriminant analysis, feature analysis, color, shape and geometric features using discrimination analysis, pulsed-phase thermography |
| 16. | Crop hail damage | Multi-spectral imaging, polarimetric radar imagery | Linear and exponential regression analysis, unsupervised image classification |
| 17. | Agricultural expansion and intensification | Satellite remote sensing in general | Wavelet-based filtering |
| 18. | Greenhouse monitoring | Optical and thermal cameras | Linear and exponential regression analysis, unsupervised classification, NDVI, IR thermography |

Appendix II: Applications of deep learning in agriculture.

| No. | Agri Area | Problem Description | Data Used | Classes and Labels | Variation among Classes | DL Model Used | FW Used | Data Pre-Processing | Data augmentation | Data for Training vs. Testing | Performance Metric Used | Value of Metric Used | Comparison with other technique | Ref. |
|---|---|---|---|---|---|---|---|---|---|---|---|---|---|---|
| 1. | Leaf classification | Classify leaves of different plant species | Flavia dataset, consisting of 1,907 leaf images of 32 species with at least 50 images per species and at most 77 images. | 32 classes: 32 Different plant species | N/A | Author-defined CNN + RF classifier | Caffe | Feature extraction based on Histograms of Curvature over Scale (HoCS), shape and statistical features, use of normalized excessive green (NExG) vegetative index, white border doubling image size, segmentation | N/A | Same. (condition variations applied in testing: translations, scaling, rotations, shading and occlusions) | CA | 97.3% ±0.6% | Feature extraction (shape and statistical features) and RF classifier (91.2% ± 1.6%) | (Hall, McCool, Dayoub, Sunderhauf, & Upcroft, 2015) |
| 2. | Leaf disease detection | 13 different types of plant diseases out of healthy leaves | Authors-created database containing 4,483 images. | 15 classes: Plant diseases (13), healthy leaves (1) and background images (1) | N/A | CaffeNet CNN | Caffe | Cropping, square around the leaves to highlight region of interest, resized to 256×256 pix, dupl. image removal | Affine transform (translation, rotation), perspective transform, and image rotations. | Same | CA | 96.30% | Better results than SVM (no more details) | (Sladojevic, Arsenovic, Anderla, Culibrk, & Stefanovic, 2016) |

| # | Category | Objective | Dataset | Classes | | Model | Framework | Preprocessing | Augmentation | Testing | Metric | Result | Notes | Reference |
|---|---|---|---|---|---|---|---|---|---|---|---|---|---|---|
| 3. | Plant disease detection | Identify 14 crop species and 26 diseases | PlantVillage public dataset of 54,306 images of diseased and healthy plant leaves collected under controlled conditions. | 38 class labels as crop-disease pairs | N/A | AlexNet, GoogleNet CNNs | Caffe | Resized to 256×256 pix., segmentation, background Information removal, fixed color casts | N/A | Same. Also tested on a dataset of downloaded images from Bing Image Search and IPM Images | F1 | 0.9935 | Substantial margin in standard benchmarks with approaches using hand-engineered features | (Mohanty, Hughes, & Salathé, 2016) |
| 4. | Plant disease detection | Classify banana leaves' diseases | Dataset of 3,700 images of banana diseases obtained from the PlantVillage dataset. | 3 classes: healthy, black sigatoka and black speckle | N/A | LeNet CNN | deeplearning4j | Resized to 60×60 pix., converted to grayscale | N/A | Same | CA, F1 | 96+% (CA), 0.968 (F1) | Methods using hand-crafted features not generalize well | (Amara, Bouaziz, & Algergawy, 2017) |
| 5. | Land cover classification | Identify 13 different land-cover classes in KSC and 9 different classes in Pavia | A mixed vegetation site over Kennedy Space Center (KSC), FL, USA (Dataset 1), and an urban site over the city of Pavia, Italy (Dataset 2). Hyperspectral datasets. | 13 different land-cover classes (Dataset 1), 9 land cover classes (Dataset 2): Soil, meadow, water, shadows, different materials | N/A | Hybrid of PCA, autoencoder (AE), and logistic regression | Developed by the authors | Some bands removed due to noise | N/A | Same | CA | 98.70% | 1% more precise than RBF-SVM | (Chen, Lin, Zhao, Wang, & Gu, 2014) |
| 6. | Land cover classification | Identify 21 land-use classes containing a variety of spatial patterns | UC Merced land-use data set. Aerial ortho-imagery with a 0.3048-m pixel resolution. Dataset compiled from a selection of 100 images/class. | 21 land-use classes: Agricultural, airplane, sports, beach, buildings, residential, forest, freeway, harbor, parking lot, river etc. | High relevance between medium density and dense residential, as well as between buildings and storage tanks | Author-defined CNN + multiview model averaging | Theano | From RGB to HSV (hue-saturation-value) color model, resized to 96×96 pix., creation of multiscale views | Views flipped horizontally or vertically with a probability of 0.5 | Same | CA | 93.48% | Unsupervised feature learning (**UFL**): 82-90% **SIFT**: 85% | (Luus, Salmon, van den Bergh, & Maharaj, 2015) |

| # | | Objective | Dataset | Classes | Confusion notes | Method | Framework | Preprocessing | | Training | Metric | Result | Comparison | Reference |
|---|---|---|---|---|---|---|---|---|---|---|---|---|---|---|
| 7. | | Extract information about cultivated land | Images from UAV at the areas Pengzhou County and Guanghan County, Sichuan Province, China. | 2 classes: Cultivated vs. non-cultivated | The cultivated land samples and part of forest land samples were easily confused | Author-defined CNN | N/A | Orthorectification, image matching, linear land elimination, correct distortion, zoomed to 40×40 pix. | N/A | Same | CA | 88-91% | N/A | (Lu, et al., 2017) |
| 8. | | Land cover classification considering time series | First dataset generated using a time series of Pléiades VHSR images at THAU Basin. Second dataset generated from an annual time series of 23 Landsat 8 images acquired in 2014 above Reunion Island. | 11 classes (dataset 1), 9 classes (dataset 2). Land cover classes such as trees, crops, forests, water, soils, urban areas, grasslands, etc. (Image object or pixel) | Tree Crops, Summer crops and Truck Farming were classes highly mixed | One-unit LSTM + RFF, One-unit LSTM + SVM | Keras/ Theano | Multiresolution segmentation technique, feature extraction, pixel-wise multi-temporal linear interpolation, various radiometric indices calculated | N/A | Same | CA, F1 | First Dataset: 75.34% (CA), 0.7463 (F1) Second Dataset: 84.61% (CA), 0.8441 (F1) | **RF and SVM** (best of both): First Dataset: 74.20% (CA), 0.7158 (F1) Second Dataset: 83.82% (CA), 0.8274 (F1) | (Ienco, Gaetano, Dupaquier, & Maurel, 2017) |
| 9. | Crop type classification | Classification of crops wheat, maize, soybeans sunflower and sugar beet | 19 multi-temporal scenes acquired by Landsat-8 and Sentinel-1A RS satellites from a test site in Ukraine. | 11 classes: water, forest, grassland, bare land, wheat, maize, rapeseed, cereals, sugar beet, sunflowers and soybeans. | General confusion between maize and soybeans | Author-defined CNN | Developed by the authors | Calibration, multi-looking, speckle filtering (3×3 window with Refined Lee algorithm), terrain correction, segmentation, restoration of missing data | N/A | Same | CA | 94.60% | **Multilayer perceptron:** 92.7%, **RF:** 88% | (Kussul, Lavreniuk, Skakun, & Shelestov, 2017) |

| # | | | | | | | | | | | | | | |
|---|---|---|---|---|---|---|---|---|---|---|---|---|---|---|
| 10. | | Classification of crops oil radish, barley, seeded grass, weed and stump | 36 plots at Foulum Research Center, Denmark containing oil radish as a catch crop and amounts of barley, grass, weed and stump. 352 patches in total. | 7 classes: oil radish, barley, weed, stump, soil, equipment and unknown (pixel of the image) | Coarse features (radish leafs and soil) were predicted quite well. Finer features (barley, grass or stump) not so much. | Adapted version of VGG16 CNN | Developed by the authors | Resized to 1600x1600 pix. centered on the sample areas, division into 400x400 pix. patches | Rotations 0, 90, 180 and 270 degrees, flipped diagonally and same set of rotations | Same | CA, IoU | 79% (CA), 0.66 (IoU) | N/A | (Mortensen, Dyrmann, Karstoft, Jørgensen, & Gislum, 2016) |
| 11. | | Crop type classification considering time series | A raster dataset of 26 SENTINEL 2A images, acquired between 2015 2016 at Munich Germany. Shortwave infrared 1 and 2 bands were selected. | 19 classes: corn, meadow, asparagus, rape, hop, summer oats, winter spelt, fallow, wheat, triticale, barley, winter rye, beans and others | Some classes represent distinct cultivated crops, others (such as meadow, fallow, triticale, wheat, and rye) are botanically related. | Three-unit LSTM | TensorFlow | Atmospherically corrected | N/A | Same | CA, F1 | 76.2% (CA), 0.558 (F1) | **CNN**: 59.9% (CA), 0.236 (F1) **SVM**: 31.7 (CA), 84.8% 0.317 (F1) | (Rußwurm & Körner, 2017) |
| 12. | | Crop type classification from UAV imagery | Aerial images of experimental farm fields issued from a series of experiments conducted by the Swiss Confederation's Agroscope research center. | 23 classes: 22 different crops plus soil (pixel of the image) | Lin and Simplex have very similar histograms | CNN + HistNN (using RGB histograms) | Keras | Image segmentation | N/A | Same | F1 | 0.90 (experiment 0), 0.73 (experiment 1) | **CNN**: 0.83 (experiment 0), 0.70 (experiment 1) **HistNN**: 0.86 (experiment 0), 0.71 (experiment 1) | (Rebetez, J., et al., 2016) |
| 13.. | Plant recognition | Recognize 7 views of different plants: entire plant, branch, flower, fruit, | LifeCLEF 2015 plant dataset, which has 91,759 images distributed in 13,887 plant observations. Each observation captures the | 1,000 classes: Species that include trees, herbs, and ferns, among others. | Images of flowers and leaf scans offer higher accuracy than the rest of the views | AlexNet CNN | Caffe | N/A | N/A | Same | LC | 48.60% | 20% worse than local descriptors to represent images and KNN, dense SIFT and a Gaussian | (Reyes, Caicedo, & Camargo, 2015) |
| | leaf, stem and scans | appearance of the plant from various points of view: entire plant, leaf branch, fruit, stem scan, flower. | | | | | | | | | | | Mixture Model | |
| 14. | Root and shoot feature identification and localisation | The first dataset contains 2,500 annotated images of whole root systems. The second hand-annotated 1,664 images of wheat plants, labeling leaf tips, leaf bases, ear tips, and ear bases. | 2 classes: Prediction if a root tip is present or not (first dataset) 5 classes: Leaf tips and bases, ear tips and bases, and negative (second dataset) | N/A | Author-defined CNN | Caffe | Image cropping at annotated locations 128x128 pix., resized to 64x64 for use in the network | N/A | Same | CA | 98.4% (first dataset) 97.3% (second dataset) | Sparse coding approach using SIFT + SVM: 80-90% | (Pound, M. P., et al., 2016) |
| 15. | Recognize 44 different plant species | MalayaKew (MK) Leaf Dataset which consists of 44 classes, collected at the Royal Botanic Gardens, Kew, England. | 44 classes: Species such as acutissima, macranthera, rubra, robur f. purpurascens etc. | N/A | AlexNet CNN | Caffe | Foreground pixels extracted using HSV color space, image cropping within leaf area | Rotation in 7 different orientations | Same | CA | 99.60% | **SVM**: 95.1%, **ANN**: 58% | (Lee, Chan, Wilkin, & Remagnino, 2015) |
| 16. | Identify plants from leaf vein patterns of white, soya and red beans | 866 leaf images provided by INTA Argentina. Dataset divided into three classes: 422 images correspond to soybean leaves, 272 to red bean leaves and 172 to white bean leaves. | 3 classes: Legume species white bean, red bean and soybean | At soybean, informative regions are in the central vein. For white and red bean, outer and smaller veins are also relevant. | Author-defined CNN | Pylearn2 | Vein segmentation, central patch extraction | N/A | Same | CA | 96.90% | Penalized Discriminant Analysis (PDA): 95.1% SVM and RF slightly worse | (Grinblat, Uzal, Larese, & Granitto, 2016) |

| # | Category | Objective | Dataset | Classes | Challenges | Model | Framework | Technique | Preprocessing | Train/Test | Metric | Results | Comparison | Reference |
|---|---|---|---|---|---|---|---|---|---|---|---|---|---|---|
| 17. | | Classify phenological stages of several types of plants purely based on the visual data | Dataset collected through TARBIL Agro-informatics Research Center of ITU, for which over a thousand agrostations are placed throughout Turkey. Different images of various plants, at different phenological stages. | 9 classes: Different growth stages of plants, starting from plowing to cropping, for the plants wheat, barley, lentil, cotton, pepper and corn. (image segment) | Appearances change very gradually and it is challenging to distinguish images falling into the growing durations that are in the middle of two successive stages. Some plants from different classes have similar color and texture distributions | AlexNet CNN | Developed by the authors | Image segmentation | Images are divided into large patches and features are extracted for each patch. 227x227 pix. patches are carved from the original images | Same | CA, F1 | 73.76 – 87.14 (CA), 0.7417 – 0.8728 (F1) | Hand crafted feature descriptors (GLCM and HOG) through a Naïve-Bayes classifier: 68.97 – 82.41 (CA), 0.6931 – 0.8226 (F1) | (Yalcin, 2017 ) |
| 18. | Plant phenology recognition | Classify the phenotyping of Arabidopsis in four accessions | Dataset composed of sequences of images captured from the plants in different days while they grow, successive top-view images of different accessions of Arabidopsis thaliana. | 4 classes: 4 different accessions of Arabidopsis: Genotype states SF-2, CVI, Landsberg (Ler) and Columbia (Col) | Plants change in size rapidly during their growth, the decomposed images from the plant sequences are not sufficiently consistent | CNN+ LSTM | Keras/ Theano | Camera distortion removal, color correction, temporal matching, plant segmentation through the GrabCut algorithm | Image rotations by 90, 180 and 270 degrees around its center | Same | CA | 93% | **Hand crafted feature descriptors + LSTM**: 68% **CNN**: 76.8% | (Namin, Esmaeilzadeh, Najafi, Brown, & Borevitz, 2017) |
| 19. | Segmentation of root and soil | Identify roots from soils | Soil images coming from X-ray tomography. | 2 classes: Root or soil (pixel of the image) | Soil/root contrast is sometimes very low | Author-defined CNN with SVM for classification | MatConvNet | Image segmentation | Simulated roots added to soil images | Same | QM | 0.23 (simulation) 0.57 (real roots) | N/A | (Douarre, Schielein, Frindel, Gerth, & Rousseau, 2016) |
| 20. | | Estimate corn yield of county level in U.S. | Corn yields from 2001 to 2010 in Illinois U.S., downloaded from Climate Research Unit (CRU), plus MODIS Enhanced Vegetation Index. | Crop yield index (scalar value) | N/A | Author-defined CNN | Caffe | Enhanced Vegetation Index (EVI), hard threshold algorithm, Wavelet transformation for detecting crop phenology | N/A | Same | RMSE | 6.298 | Support Vector Regression (SVR): 8.204 | (Kuwata & Shibasaki, 2015) |
| 21. | Crop yield estimation | Mapping winter vegetation quality coverage considering time series | Sentinel-1 dataset including 13 acquisitions in TOPS mode from October 2016 to February 2017, with a temporal baseline of 12 days. Dual-polarization (VV+VH) data in 26 images. | 5 classes: Estimations of the quality of vegetative development as bare soil, very low, low, average, high | "Low" class intersects the temporal profiles of all the other classes multiple times. A misclassification rate exists between the "low" and "bare soil" classes | Five-unit LSTM, Gated Recurrent Unit (GRU) | Keras/ Theano | Intensity image gen., radiometrical calibration, temporal filtering for noise reduction, orthorectification into map coordinates, transformed to logarithm scale, normalized | N/A | Same | CA, F1 | 99.05% (CA), 0.99 (F1) | **RF and SVM** (best of both): 91.77% (CA), 0.9179 (F1) | (Minh, et al., 2017) |
| 22. | Fruit counting | Predict number of tomatoes in the images | 24,000 synthetic images produced by the authors. | Estimated number of tomato fruits (scalar value) | N/A | Modified Inception-ResNet CNN | TensorFlow | Blurred synthetic images by a Gaussian filter | Generated synthetic 128x128 pix. images to train the network, colored circles to simulate background and tomato plant/crops. | Trained entirely on synthetic data and tested on real data | RFC, RMSE | 91% (RFC) 1.16 (RMSE) on real images, 93% (RFC) 2.52 (RMSE) on synthetic images | ABT: 66.16% (RFC), 13.56 (RMSE) | (Rahmenoonfar & Sheppard, 2017) |

| # | | Problem | Dataset | Output | Variation | Model | Framework | Annotation | Augmentation | Training/Testing | Metric | Result | Comparison | Reference |
|---|---|---|---|---|---|---|---|---|---|---|---|---|---|---|
| 23. | | Map from input images of apples and oranges to total fruit counts | 71 1280×960 orange images (day time) and 21 1920×1200 apple images (night time). | Number of orange or apple fruits (scalar value) | High variation in CA. For orange, dataset has high occlusion, depth variation, and uncontrolled illumination. For apples, data set has high color similarity between fruit/foliage | CNN (blob detection and counting) + Linear Regression | Caffe | Image segmentation for easier data annotation by users, creation of bounding boxes around image blobs | Training set partitioned into 100 randomly cropped and flipped 320×240 pix. sub-images | Same (but different trees used in training and testing) | RFC, L2 | 0.968 (RFC), 13.8 (L2) for oranges 0.913 (RFC), 10.5 (L2) for apples | Best texture-based regression model: 0.682 (RFC) | (Chen, et al., 2017) |
| 24. | | Fruit detection in orchards, including mangoes, almonds and apples | Images of three fruit varieties: apples (726), almonds (385) and mangoes (1,154), captured at orchards in Victoria and Queensland, Australia. | Sections of apples, almonds and mangoes at the image (bounding box) | Within class variations due to distance to fruit illumination, fruit clustering, and camera view-point. Almonds similar in color and texture to the foliage | Faster Region-based CNN with VGG16 model | Caffe | Image segmentation for easier data annotation | Flip, scale, flip-scale and the PCA augmentation technique presented in AlexNet | Same | F1-IoU (20) | 0.904 (apples) 0.908 (mango) 0.775 (almonds) | ZF network: 0.892 (apples) 0.876 (mango) 0.726 (almonds) | (Bargoti & Underwood, 2016) |
| 25. | | Detection of sweet pepper and rock melon fruits | 122 images obtained from two modalities: color (RGB) and Near-Infrared (NIR). | Sections of sweet red peppers and rock melons on the image (bounding box) | Variations to camera setup, time and locations of data acquisition. Time for data collection is day and night, sites are different. Varied fruit ripeness. | Faster Region-based CNN with VGG16 model | Caffe | Early/late fusion techniques for combining the classification info from color and NIR imagery, bounding box segmentation, pairwise IoU | N/A | Same (authors demonstrate by using a small dataset that the model can generalize) | F1-IoU (40) | 0.838 | Conditional Random Field to model color and visual texture features: 0.807 | (Sa, et al., 2016) |
| 26. | Obstacle detection | Identify ISO barrel-shaped obstacles in | 437 images from authors' experiments and recordings, 1,925 positive and | Identify if a barrel-shaped object is present in the image | N/A | AlexNet CNN | Caffe | Resized to 114×114 pix., bounding boxes of the object created | Various rotations at 13 scales, intensity of the object | Testing in different fields (row crops, grass mowing), | CA-IoU (50) | 99.9% in row crops and 90.8% in | N/A | (Steen, Christiansen, Karstoft, & |

| No. | Category | Task | Dataset | Classes | Similarity | Model | Framework | Preprocessing | Augmentation | Same | Metric | Result | Comparison | Reference |
|---|---|---|---|---|---|---|---|---|---|---|---|---|---|---|
| | | row crops and grass mowing | 11,550 negative samples. | (bounding box) | | | | | adapted | containing other obstacles (people and animals) | grass mowing | | | Jørgensen, 2016) |
| 27. | | Detect obstacles that are distant, heavily occluded and unknown | Background data of 48 images and test data of 48 images from annotations of humans, houses, barrels, wells and mannequins. | Classify each pixel as either foreground (contains a human) or background (anomaly detection) | N/A | AlexNet and VGG CNNs | Caffe | Image cropping, resized by a factor of 0.75 | N/A | Same | F1-IoU (50) | 0.72 | Local de-correlated channel features: 0.113 | (Christiansen, Nielsen, Steen, Jørgensen, & Karstoft, 2016) |
| 28. | | Classify 91 weed seed types | Dataset of 3,980 images containing 91 types of weed seeds. | 91 classes: Different common weeds found in agricultural fields | Similarity between some classes is very high (only slight differences in shape, texture, and color) | PCANet + LMC classifiers | Developed by the authors | Image filter extraction through PCA filters bank, binarization and histograms' counting | N/A | Same (also scaling for a certain range translation distance and rotation angle | CA | 90.96% | Manual feature extraction techniques + LMC classifiers: 64.80% | (Xinshao & Cheng, 2015) |
| 29. | Identification of weeds | Classify weed from crop species based on 22 different species in total. | Dataset of 10,413 images, taken mainly from BBCH 12-16 containing 22 weed and crop species at early growth stages. | 22 classes: Different species of weeds and crops at early growth stages e.g. chamomile, knotweed, cranesbill, chickweed and veronica | Variations with respect to lighting, resolution, and soil type. Some species (Veronica, Field Pancy) were very similar and difficult to classify | Variation of VGG16 | Theano-based Lasagne library for Python | Green segmentation to detect green pixels, non-green pixels removal, padding added to make images square, resized to 128x128 pix. | Image mirroring and rotation in 90 degree increments | Same | CA | 86.2% | Local shape and color features: 42.5% and 12.2% respectively | (Dyrmann, Karstoft, & Midtiby, 2016 ) |
| 30. | | Identify thistle in | 4,500 images from 10, 20, 30, and 50m of altitude | 2 classes: Whether the image contains | Small variations in some images depending on | DenseNet CNN | Caffe | Image cropping | Random flip both horizontally | Same (extra tests for the case of | CA | 97% | Color feature-based Thistle-Tool: 95% | (Sørensen, Rasmuss |

| | | winter wheat and spring barley images | captured by a Canon PowerShot G15 camera. | thistle in winter wheat or not (Heatmap of classes is generated at the output) | the percentage of thistles they contain | | | and vertically, random transposing | winter barley) | | | | en, Nielsen, & Jørgensen, 2017) |
|---|---|---|---|---|---|---|---|---|---|---|---|---|---|
| 31. | | Weed segmentation for robotic platforms | Crop/Weed Field Image Dataset (CW-FID), consists of 20 training and 40 testing images. A dataset of 60 top-down field images of a common culture (organic carrots) with the presence of intra-row and close-to-crop weeds. | 2 classes: carrot plants and weeds (image region) | N/A | Adapted version of Inception-v3 + lightweight DCNN + set of K lightweight models as a mixture model (MixDCNN) | TensorFlow | Image up-sampling to 299x299 pix., NDVI-based vegetation masks, extracting regions based on a sliding window on the color image | N/A | Same (different carrot fields used for testing) | CA | 93.90% | Feature extraction (shape and statistical features) and RF classifier: 85.9% | (McCool, Perez, & Upcroft, 2017) |
| 32. | | Automating weed detection in color images despite heavy leaf occlusion | 1,427 images from winter wheat fields, of which 18,541 weeds have been annotated, collected using a camera mounted on an all-terrain vehicle. | Detect single weed instances in images of cereal fields (bounding box). A coverage map is produced. | Large parts of the weeds overlap with wheat plants | Based on DetectNet CNN (which is based on GoogLeNet CNN) | Developed by the authors | Resized to 1224×1024 pix. | N/A | Different field used for testing. This field has a severe degree of occlusion compared to the others | IoU P- IoU (50) R-IoU (50) | 0.64 (IoU), 86.6% (P- IoU), 46.3% (R-IoU) | N/A | (Dyrmann, Jørgensen, & Midtiby, 2017) |
| 33. | Crop/weed detection and classification | Detecting sugar beet plants and weeds in the field based on image data | 1,969 RGB+NIR images captured using a JAI camera in nadir view placed on a UAV. | Identify if an image patch belongs to weed or sugar beet (image region) | N/A | Author-defined CNN | TensorFlow | Separated vegetation/ background based on NDVI, binary mask to describe vegetation, blob segmentation, resized to 64x64 pix., normalized and centered | 64 even rotations | Same (also generalized to a second dataset produced 2-weeks after, at a more advanced growth stage) | P, R | Dataset A: 97% (P), 98% (R) Dataset B: 99% (P), 89% (R) | N/A | (Milioto, Lottes, & Stachniss, 2017) |
| 34. | | Detecting and classifying sugar beet plants and weeds | 1,600 4-channels RGB+NIR images captured before (700 images) and after (900 images) a 4-week period, provided by a multispectral JAI camera mounted on a BOSCH Bonirob farm robot. | Identifies if a blob belongs to sugar beet crop, weeds or soil (image blob) | N/A | Author-defined CNN | TensorFlow | Pixel-wise segmentation between green vegetation and soil based on NDVI and light CNN, unsupervised dataset summariz. | N/A | Same (also generalized to a second dataset produced 4-weeks after, at a more advanced growth stage) | CA | 98% (Dataset A), 59.4% (Dataset B) | Feature extraction (shape and statistical features) and RF classifier: 95% | (Potena, Nardi, & Pretto, 2016) |
| 35. | | Detecting and classifying weeds and maize in fields | Simulated top-down images of overlapping plants on soil background A total of 301 images of soil and 8,430 images of segmented plants. The plants cover 23 different weed species and maize. | Identifies if an image patch belongs to weed, soil or maize crop (image pixel) | N/A | Adapted version of VGG16 CNN | Developed by the authors | Image cropping in 800x800 pix. | Random scaling from 80 to 100% of original size, random rotations in one degree increments, varied hue, saturation and intensity, random shadows | Tested on real images while trained on simulated ones | CA, IoU | 94% CA, 0.71 IoU (crops), 0.70 IoU (weeds) 0.93 IoU (soil) | N/A | (Dyrmann, Mortensen, Midtiby, & Jørgensen, 2016) |
| 36. | Prediction of soil moisture content | Predict the soil moisture content over an irrigated corn field | Soil data collected from an irrigated corn field (an area of 22 sq. km) in the Zhangye oasis, Northwest China. | Percentage of soil moisture content (SMC) (scalar value) | N/A | Deep belief network-based macroscopic cellular automata (DBN-MCA) | Developed by the authors | Geospatial interpolation for creation of soil moisture content maps, multivariate geostatistical approach for estimating thematic soil maps, maps converted to TIFF, resampled to 10-m res. | N/A | Same | RMSE | 6.77 | Multi-layer perceptron MCA (MLP-MCA): 18% reduction in RMSE | (Song, et al., 2016) |

| | | | | | | | | | | | | |
|---|---|---|---|---|---|---|---|---|---|---|---|---|
| 37. | | Practical and accurate cattle identification from 5 different races | 1,300 images collected by the authors. | 5 classes: Cattle races, Bali Onggole or Pasuruan, Aceh Madura and Pesisir | N/A | GLCM – CNN | Deep Learning Matlab Toolbox | GLCM features extraction (contrast, energy and homogeneity), saliency maps to accelerate feature extraction | N/A | Same | CA | 93.76% | **CNN** without extra inputs: 89.68% **Gaussian Mixture Model** (GMM): 90% | (Santoni, Sensuse, Arymurth y, & Fanany, 2015) |
| 38. | Animal research | Predict growth of pigs | 160 pigs, housed in two climate controlled rooms, four pens/room, 10 pigs/pen. Ammonia, ambient and indoor air temperature and humidity, feed dosage and ventilation measured at 6-minute intervals. | Estimation of the weight of pigs (scalar value) | N/A | First-order DRNN | Develop ed by the authors | N/A | N/A | Tested on different rooms of pigs than the ones which were used for training | MSE, MRE | 0.002 (MSE) on same dataset), 10% (MRE) in relation to a controller | N/A | (Demmer s T. G., Cao, Parsons, Gauss, & Wathes, 2012) |
| 39. | | Control of the growth of broiler chickens | Collecting data from 8 rooms, each room housing 262 broilers, measuring bird weight, feed amount, light intensity and relative humidity. | Estimation of the weight of chicken (scalar value) | N/A | First-order DRNN | Develop ed by the authors | N/A | N/A | Tested on different rooms of chicken than the ones which were used for training | MSE, MRE | 0.02 (MSE), 1.8% (MRE) in relation to a controller | N/A | (Demmer s T. G., et al., 2010) |
| 40. | Weather prediction | Predict weather based on previous year's conditions | Syngenta Crop Challenge 2016 dataset, containing 6,490 sub-regions with three weather condition attributes from the years 2000 to 2015. | Predicted values of temperature, precipitation and solar radiation (scalar value) | N/A | LSTM | Keras | N/A | N/A | Same | N-RMSE, MRE | 78% (Temperature), 73% (Precipitation), 2.8% (Solar Radiation) N-RMSE, 1-3% MRE in all categories | N/A | (Sehgal, et al., 2017) |

1
2

Appendix III: Publicly-available datasets related to agriculture.

| No. | Organization/Dataset | Description of dataset | Source |
|-----|---------------------|------------------------|--------|
| 1. | Image-Net Dataset | Images of various plants (trees, vegetables, flowers) | http://image-net.org/explore?wnid=n07707451 |
| 2. | ImageNet Large Scale Visual Recognition Challenge (ILSVRC) | Images that allow object localization and detection | http://image-net.org/challenges/LSVRC/2017/#det |
| 3. | University of Arcansas, Plants Dataset | Herbicide injury image database | https://plants.uaex.edu/herbicide/ <br> http://www.uaex.edu/yard-garden/resource-library/diseases/ |
| 4. | EPFL, Plant Village Dataset | Images of various crops and their diseases | https://www.plantvillage.org/en/crops |
| 5. | Leafsnap Dataset | Leaves from 185 tree species from the Northeastern United States. | http://leafsnap.com/dataset/ |
| 6. | LifeCLEF Dataset | Identity, geographic distribution and uses of plants | http://www.imageclef.org/2014/lifeclef/plant |
| 7. | PASCAL Visual Object Classes Dataset | Images of various animals (birds, cats, cows, dogs, horses, sheep etc.) | http://host.robots.ox.ac.uk/pascal/VOC/ |
| 8. | Africa Soil Information Service (AFSIS) dataset | Continent-wide digital soil maps for sub-Saharan Africa | http://africasoils.net/services/data/ |
| 9. | UC Merced Land Use Dataset | A 21 class land use image dataset | http://vision.ucmerced.edu/datasets/landuse.html |
| 10. | MalayaKew Dataset | Scan-like images of leaves from 44 species classes. | http://web.fsktm.um.edu.my/~cschan/downloads_MKLeaf_dataset.html |
| 11. | Crop/Weed Field Image Dataset | Field images, vegetation segmentation masks and crop/weed plant type annotations. | https://github.com/cwfid/dataset <br> https://pdfs.semanticscholar.org/58a0/9b1351ddb447e6abded e7233a4794d538155.pdf |
| 12. | University of Bonn Photogrammetry, IGG | Sugar beets dataset for plant classification as well as localization and mapping. | http://www.ipb.uni-bonn.de/data/ |
| 13. | Flavia leaf dataset | Leaf images of 32 plants. | http://flavia.sourceforge.net/ |
| 14. | Syngenta Crop Challenge 2017 | 2,267 of corn hybrids in 2,122 of locations between 2008 and 2016, together with weather and soil conditions | https://www.ideaconnection.com/syngenta-crop-challenge/challenge.php |